\documentclass{article}

\usepackage[preprint,nonatbib]{neurips_2020}

\usepackage[utf8]{inputenc} % allow utf-8 input
\usepackage[T1]{fontenc}    % use 8-bit T1 fonts
\usepackage{hyperref}       % hyperlinks
\usepackage{url}            % simple URL typesetting
\usepackage{booktabs}       % professional-quality tables
\usepackage{amsfonts}       % blackboard math symbols
\usepackage{nicefrac}       % compact symbols for 1/2, etc.
\usepackage{microtype}      % microtypography
\usepackage{breqn}
\usepackage{color}
\usepackage{pdfpages}
\usepackage[style=base]{caption}
\usepackage{subfig}
\usepackage{graphicx}

\title{Improving Stability of LS-GANs for Audio and Speech Signals}

\author{
Mohammad Esmaeilpour, Raymel Alfonso Sallo, Olivier St-Georges\\
\bf Patrick Cardinal, Alessandro Lameiras Koerich\\
\'{E}cole de Technologie Sup\'{e}rieure (\'{E}TS)\\
Universit\'{e} du Qu\'{e}bec\\
Montreal, QC, Canada \\
}

\begin{document}

\maketitle

\begin{abstract}
In this paper we address the instability issue of generative adversarial network (GAN) by proposing a new similarity metric in unitary space of Schur decomposition for 2D representations of audio and speech signals. We show that encoding departure from normality computed in this vector space into the generator optimization formulation helps to craft more comprehensive spectrograms. We demonstrate the effectiveness of binding this metric for enhancing stability in training with less mode collapse compared to baseline GANs. Experimental results on subsets of UrbanSound8k and Mozilla common voice datasets have shown considerable improvements on the quality of the generated samples measured by the Fr\'{e}chet inception distance. Moreover, reconstructed signals from these samples, have achieved higher signal to noise ratio compared to regular LS-GANs.
\end{abstract}

%%%%%%%%%%%%%%%%%%%%%%%%%%%%%%%%%%%%
\section{Introduction}
%%%%%%%%%%%%%%%%%%%%%%%%%%%%%%%%%%%%
Generative models have been widely used in several audio and speech processing tasks such as verification \cite{reynolds2000speaker}, enhancement \cite{chehrehsa2016speech}, synthesis \cite{raitio2010hmm}, etc. In the last few years, generative adversarial networks (GANs) \cite{goodfellow2014generative} have much contributed for tackling such challenging tasks and many GAN architectures have been introduced \cite{bollepalli2019generative, sriram2018robust}. Furthermore, GANs have been employed in high-level data augmentation approaches for supervised, semi-supervised, and unsupervised audio and speech classification tasks \cite{hu2018generative, donahue2018exploring}. In augmentation with paired transformations, cycle-consistent GANs have been developed for environmental sound classification \cite{esmaeilpour2020unsupervised} as well as for more sophisticated tasks such as voice conversion \cite{fang2018high}. A typical GAN optimizes the minimax problem between two networks of generator ($G$) and discriminator ($D$), where the latter should discriminate the real sample distribution ($p_{r}$) from the generated sample distribution ($p_{g}$) using the Jensen–Shannon divergence (JSD) similarity metric. Following such a baseline GAN, many other variants of GANs have been introduced, which utilize updated similarity measures such as the Wasserstein metric \cite{arjovsky2017wasserstein} and the least squares loss \cite{mao2017least}. However, training and tuning such GAN models for audio and speech processing tasks have always been a mounting concern and a difficult challenge due to their instability, mode collapse, and oversmoothing issues \cite{esmaeilpour2020unsupervised}. Besides that, there are several other issues on training GANs with the gradient descent procedure such as the difficulty in reaching Nash equilibrium between $D$ and $G$ \cite{salimans2016improved}, the instability limitation is due to lying in low dimensional manifolds of $p_{r}$ and $p_{g}$ \cite{arjovsky2017towards}, weakness of vanilla loss function formulation for $D$ \cite{yang2017statistical}, among others.

In this paper, we address stability and mode collapse issues of advanced GANs by encoding a new similarity metric defined in the vector space of Schur decomposition \cite{van1983matrix, esmaeilpour2020detection} as a part of the generator optimization formulation. Our main contribution in this paper is proposing GAN configurations for 2D representations of audio and speech signals for improving stability and reducing mode collapse, which are typical issues with GANs. The experimental results on environmental sound and voice common datasets have shown least square GAN encoded with this metric outperforms baseline GANs and produces higher quality samples. 

The organization of this paper is as follows. In Section~\ref{sec:rw}, we review some related works and in Section~\ref{sec:fgan}, we study properties of GANs in greater details. Section~\ref{sec:dep} presents theoretical concepts for developing a new similarity metric in the unitary vector space. We provide our experimental results and associated discussions in Section~\ref{exp:sec}. Finally, the conclusion and the perspectives of future work are presented.

%%%%%%%%%%%%%%%%%%%%%%%%%%%%%%%%%%%%
\section{Related Works}
\label{sec:rw}
%%%%%%%%%%%%%%%%%%%%%%%%%%%%%%%%%%%%
The vanilla GAN employs the JSD symmetric similarity metric between two distributions without benefiting neither Markov chain nor approximate inference models which is common for generative models (e.g., hidden Markov models). The generator network $G(\bold{z};\theta_{g})$ learns to map from the latent space and the random noise distribution $p_{z}$ to the generated distribution $p_{g}$. The discriminator network $D(\bold{x}; \theta_{r})$ learns to maximize $\mathbb{E}_{\bold{z}\sim p_{z}(\bold{z})}\left [ \log \left ( 1-D(G(\bold{z})) \right ) \right ]$ against $G$ as defined in (\ref{gan1}).
\begin{equation}
    \mathbb{E}_{\bold{x}\sim p_{r}(\bold{x})}\left [ \log D(\bold{x}) \right ]+\mathbb{E}_{\bold{x}\sim p_{g}(\bold{x})}\left [ \log \left ( 1-D(G(\bold{x})) \right ) \right ]
    \label{gan1}
\end{equation}
This minimax optimization problem theoretically should yield a generalizable discriminator with discretized parameters $\theta_{r}$ in its learning subspace. However, the optimization process can be unstable and collapse for different modes \cite{salimans2016improved, thanh2019improving}. This might happen when $D(\bold{x}; \theta_{r})$ cannot detect disjoint mode distributions in $(p_{g}, p_{r})$. Following this reasoning and for rectifying this issue, several reconstructor frameworks have been developed for mapping from $p_{g}$ to the noise distribution $p_{z}$ \cite{donahue2016adversarial, srivastava2017veegan}. The reconstructor network can be as simple as an autoencoder for keeping consistency between $p_{z}$ and $p_{g}$ minimizing for (\ref{veegan}) \cite{srivastava2017veegan}.
\begin{equation}
    \mathbb{E}\left [ \left \| \bold{z}-R(G(\bold{z})) \right \|_{2}^{2} \right ]+H(\bold{z},R(\bold{x}))
    \label{veegan}
\end{equation}
\noindent where $R$ denotes the reconstructor network (the autoencoder) and $H$ is the entropy loss with respect to the joint distribution $p_{0}(z)p_{g}(x|z)$ for $p_{0}\sim \mathcal{N}(0,1)$. In this framework, $R$ resembles a discriminator independent prior over the joint generated and real distribution subspaces which has shown to improve the quality of the generated samples. However, it may not provide a generalizable multi-mode expansion over all the generator distribution in discriminable subspace $D(\bold{x}; \theta_{r})$. Reconstructor networks with similar intuition as (\ref{veegan}) in variational encoder enhancement (VEE) framework \cite{srivastava2017veegan} have been previously configured to be dependent on the discriminator so that reducing the chance of undetectable modes and training instability; such as bidirectional GAN \cite{donahue2016adversarial} and adversarially learning inference \cite{dumoulin2016adversarially}. Partial inclusion of reconstructor network within the training process of $G(\bold{z};\theta_{g})$ has also been studied in \cite{chen2016infogan} which unlike the VEE configuration, it does not incorporate inverse translation into $p_{r}$.

In addition to the vanilla autoencoder-based reconstructors, some adversarial methods have been proposed for autoencoders aiming at reducing mode collapse side effect. The effectiveness of explicit regularization of loss functions with autoencoder loss has been studied by Che \textit{et al.}~\cite{che2016mode}. They have introduced several costly metrics for estimating missing modes and quality of the generated samples. A similar metric for autoencoding random samples from $p_{r}$ to $p_{g}$ has been introduced by Larsen \textit{et al.}~\cite{larsen2015autoencoding}. Both these approaches incorporate $\mathcal{L}\left [ \bold{x}, G(R(\bold{x})) \right ]$ which is a pixel-wise loss followed by a regularization term. Some variational autoencoder schemes have been also proposed \cite{kingma2014stochastic, rezende2014stochastic} for encoding real sample distribution rather than noise vectors so that characterizing potential mode mismatches.

Salimans \textit{et al.}~\cite{salimans2016improved} have explained the instability issue of GANs with difficulties in finding Nash equilibrium in training. This becomes more critical when the generator and the discriminator minimize cost functions (at every hidden layer) with dissimilar signs. Hence, the gradient backpropagation causes instability for both generator and discriminator networks. In order to potentially rectify this problem, they firstly diagnosed that the generator overtrains amidst the discriminator being trained. Therefore, they updated the generator ($G$) objective as \cite{salimans2016improved}:
\begin{equation}
    \left \| \mathop{{}\mathbb{E}}_{\bold{x}\sim p_{r}}f(\bold{x}) -  \mathop{{}\mathbb{E}}_{\bold{z}\sim p_{z}(\bold{z})}f(G(\bold{z}))\right \|_{2}^{2}
    \label{feature_matching}
\end{equation}
\noindent where $f(\bold{x})$ denotes the post activation function for the discriminator. This update does not apply neither to the discriminator nor $f$, but it forces the generator to produce statistical features which match real data distribution. Even with the assumption of having a constant optimal discriminator, still there is no guarantee to detect all the possible modes in $p_{r}$.

From the discriminator perspective, the generator is a density function which transforms a random noise distribution to realistic representations. This density function implicitly learns similarity (closeness) among training sample distributions. Explicit integration of intra-class similarity as a separate input vector into $D(\bold{x}; \theta_{r})$ was introduced as minibatch discrimination \cite{salimans2016improved}. This technique in some scenarios helps to avoid mode collapse featuring intra-similarities mediated by an exponential function. However, not only it does not completely characterize instability issue of networks, but also it might shatter gradient information as well.

Historical averaging \cite{salimans2016improved} is an approach based on the common regularization trick for avoiding overtraining in GANs. It modifies the loss functions of both $D(\bold{x}; \theta_{r})$ and $G(\bold{z};\theta_{g})$ in order to support balanceness in training. This modification in general should yield a stable behavior in training, but for disjoint distributions of samples in the subspace of $p_{r}$ it might fail due to utilization of JSD \cite{manning1999foundations}.

Unlike the vanilla GAN which incorporates JSD similarity metric, the Wasserstein GAN \cite{arjovsky2017wasserstein} incorporates the Wasserstein metric into its loss function so that it trains a more stable generative model. For lower-dimensional manifolds, this metric can better characterize the similarities between two distributions. However, for overlapping submanifolds with disjoint distribution, Wasserstein GAN might also be unstable because of vanishing gradient \cite{gulrajani2017improved}. The majority of the approaches introduced for discussing stability in GANs, propose loss function modifications to support a more comprehensive similarity measure between real and noisy distributions (in addition to latent parameters). The least squares generative adversarial networks (LS-GAN) \cite{mao2017least} addresses this issue by minimizing Pearson $\chi^{2}$ divergence. It has been shown that, compared to the vanilla GAN and the Wasserstein GAN, the LS-GAN can generate higher quality samples. However, it might fail to keep stability of networks ~\cite{karras2017progressive}.

While instability, mode collapse, and oversmoothing are common problems of GANs trained on any dataset, they pose more training concerns for audio and speech datasets mainly because of the complex nature of Mel-frequency cepstral coefficient, short-time Fourier transform, and discrete wavelet transform (DWT) representations \cite{esmaeilpour2020unsupervised}, which are commonly used to represent audio and speech signals. These representations produce spectrograms (images) that are fundamentally different from natural images and they might easily make the generator collapse.

%%%%%%%%%%%%%%%%%%%%%%%%%%%%%%%%%%%%
\section{f-GAN: a Brief Overview}
\label{sec:fgan}
%%%%%%%%%%%%%%%%%%%%%%%%%%%%%%%%%%%%
For maximizing the similarity between $p_{r}$ and $p_{g}$ distributions, many divergence metrics have been investigated to meaningfully characterize the latent properties of sample distributions in Cartesian and convex spaces. f-GAN \cite{nowozin2016f} implements the f-divergence metric defined in (\ref{f_divergence}) for approximating the potential dissimilarity between sample distributions \cite{hong2019generative}. 
\begin{equation}
    D_{f}(p_{r}\parallel p_{g}) = \int p_{g}(\bold{x})f\left ( \frac{p_{r}(\bold{x})}{p_{g}(\bold{x})} \right )dx
    \label{f_divergence}
\end{equation}
\noindent where in the closed-form representation, $f$ (the generator) is a convex function and $p_{g}(\bold{x})$ is distributed over all the possible random input $\bold{x}$. Encoding this equation into f-GAN in its generic representation requires an estimation for any arbitrary function $f$ using a convex conjugate format of $f(x) = \sup \left ( tx-f^{*}(t) \right )$ where $f^{*}(t)$ denotes the Fenchel conjugate for $f$ \cite{hong2019generative, fenchel1949conjugate}; and $t$ should be defined in the possible domain of $f^{*}$. With this assumption, (\ref{f_divergence}) can be parameterized as the expectation of translated distributions governed by the discriminator $D$, as shown in (\ref{f_gan1}) \cite{hong2019generative}.
\begin{equation}
   \sup \left ( \mathop{{}\mathbb{E}}_{\bold{x}\sim p_{r}}\left [ D(\bold{x}) \right ]-\mathop{{}\mathbb{E}}_{\bold{x}\sim p_{g}}\left [f^{*}(actv(D(\bold{x}))) \right ] \right )
    \label{f_gan1}
\end{equation}
\noindent where $actv$ stands for the activation function. This divergence metric is comprehensive enough to derive JSD, Kullback-Leibler (KL), and its reverse function for any minimax optimization problem. Using this framework, various GANs can be constructed with specific generator function $f(t)$, from the vanilla GAN to the energy-based generative adversarial network \cite{zhao2016energy}. Among all these advanced GANs with symmetric (or even asymmetric) divergence metrics, LS-GAN has shown better performance \cite{hong2019generative}. The optimal generator function $f(t)$ for this generative model is $(t-1)^{2}$ for values of $t$ defined in the domain of its associated Fenchel conjugate. Following these propositions, the least square loss function used in the discriminator network can be defined as (\ref{lsgna_d}).
\begin{equation}
    \min_{D}\frac{1}{2}\mathop{{}\mathbb{E}}_{\bold{x}\sim p_{r}}\left [ D(\bold{x})-b)^{2} \right ]+\mathop{{}\mathbb{E}}_{\bold{z} \sim p_{z}}\left [ (D(G(\bold{z}))-a)^{2} \right ]
    \label{lsgna_d}
\end{equation}
\noindent where $a$ and $b$ refer to designated values for real samples. Moreover, LS-GAN solves for $z \sim p_{z}$ using (\ref{lsgna_d2}) \cite{hong2019generative}:
\begin{equation}
    \min_{G}\frac{1}{2}\mathop{{}\mathbb{E}}_{\bold{z}\sim p_{z}}\left [ (D(G(\bold{z}))-c)^{2} \right ]
    \label{lsgna_d2}
\end{equation}
\noindent where $c$ is the designated value for the generated samples. Because of the simplicity and comprehensiveness of the LS-GAN, we use it as a benchmarking generative model in the current study.

%%%%%%%%%%%%%%%%%%%%%%%%%%%%%%%%%%%%
\section{Departure from Normality in Sample Distributions}
\label{sec:dep}
%%%%%%%%%%%%%%%%%%%%%%%%%%%%%%%%%%%%
In this section we provide some theoretical concepts for addressing functionality issues of GANs with special focus on 2D representations of audio and speech signals. For an input sample (spectrogram) $\bold{x}_{i}$ from a given distribution, there exists a unitary representation $Q \in \mathbb{C}^{n \times n}$ in such a way that \cite{golub2012matrix}:
\begin{equation}
    Q^{H}\bold{x}_{i}Q = V+S
\end{equation}
\noindent where $Q^H$ denotes the conjugate transpose of $Q$ in vector space of Schur decomposition and $S=\left \{ s_{i}\mid i=0:n-1 \right \} \in \mathbb{C}^{n \times n}$ is a upper triangular matrix. Moreover, $V=\mathrm{diag}(\lambda_{0},\lambda_{1},\cdots,\lambda_{n-1})$ contains eigenvalues of $\bold{x}_{i}$ ($\lambda$ is set of eigenvector for $\bold{x}_{i}$). In this unitary vector space, yielding the quasi-upper triangular representation $S$, $Q=\left [ q_{0}\mid  q_{1} \mid  q_{2} \mid  \cdots \mid  q_{n-1} \right ]$ provides the pencil of $\overrightarrow{q_{i}}-\lambda_{i} \overrightarrow{q_{i+1}}$ for $i\leq n-2$ which is also known as basis vector of $\bold{x}_{i}$. Therefore, linear combination of the input distribution $\bold{x}_{i}$ with variants of the support matrix $S$ can be approximated as shown in (\ref{basisfunc}) \cite{golub2012matrix}.
\begin{equation}
    \bold{x}_{i}q_{k}\approx \lambda_{k}q_{k}+\sum_{i=0}^{n-1}s_{ik}q_{i}, \quad k=0:n-1
    \label{basisfunc}
\end{equation}
\noindent with the assumption of having strictly upper triangular subspaces with span of $\left \{ q_{0},q_{1}, \cdots, q_{k} \right \}$ for $k=0:n-1$, and that the distribution of $S$ is independent of $Q$ \cite{golub2012matrix}. Therefore we can compute its Frobenius norm using eigenvalues $\lambda_{i}$ as \cite{golub2012matrix}:
\begin{equation}
    \left \| S \right \|_{F}^{2} = \left \| \bold{x}_{i} \right \|_{F}^{2}-\sum_{i=0}^{n-1}\left | \lambda_{i} \right |^{2}\equiv \Delta^{2} (\bold{x}_{i})
    \label{f_gan2}
\end{equation}
\noindent where $\Delta^{2}$ is known as departure from normality (DFN). For two samples $\bold{x}_{r}$ and $\bold{x}_{g}$ randomly drawn from $p_{r}$ and $p_{g}$
%\aktd{Please revise the preceding paragraph. two matching distributions??? $\bold{x}_.$ is an array, an input sample, not a distribution. $p_{.}$ is a distribution!}
% Respond: You are right. I edited it.
in their designated vector spaces (span of $q_{i}$s), the DFN metric should support $\left | \Delta^{2}(\bold{x}_{j}) - \Delta^{2}(\bold{x}_{i}) \right |<\epsilon$ for a small enough $\epsilon$, where $\epsilon\leq  \max_{i\geq 0, j\leq n}\left | \lambda_{i}\right |/{\left | \lambda_{j} \right |}$. This inequality ensures consistency of corresponding eigenvalues for the input samples and implies the generalized form of Schur decomposition (aka QZ decomposition) with $Q_{k}^{H}\bold{x}_{r}Z_{k}=R_{k}(V_{k}+S_{k})$ and $Q_{k}^{H}\bold{x}_{g}Z_{k}=R_{k}(V_{k}+S_{k})$ where \cite{golub2012matrix}:
\begin{equation}
R_{k}=Q_{k}^{H}(\bold{x}_{r}\bold{x}_{g_{k}}^{-1}Q_{k})
\end{equation}
\noindent herein, $Z_{k}$ is also unitary and supports for $\lim_{i\rightarrow \infty}(Q_{k_{i}}, Z_{k_{i}})=(Q,Z)$. This is a very important property for local distributions $\bold{x}_{r}$ and $\bold{x}_{g}$ mainly because it characterizes independency of the corresponding sample distributions to the basis vectors $Q$ and $Z$. The intuition behind exploiting these basis vectors is providing pencils of  $\overrightarrow{\bold{x}_{r_{i}}}-\lambda_{i}\overrightarrow{\bold{x}_{g_{i}}}$ for reconstructing the original distribution $p_{r}$ and the generator distribution $p_{g}$. The derivable pencils are not necessarily normal in the span of their associated subspaces, however, their linear combinations with the summation of $S$ and $V$ representations reconstruct $p_{r}$ and $p_{g}$ (in the closed form). Diagonal values in $V$ constitutes coefficient of basis vectors (pencils in their manifolds) and represent generic distribution for the given input sample. Product of the basis vectors in $Q$ and $Z$ which primarily encode latent modes (because of providing basis vectors resembling distribution characteristics over $Q$ and $Z$) of $p_{r}$ and $p_{g}$ with the derived pencils yields a proper approximation for sample distributions within an achievable scalar infimum and supremum.

\noindent \textbf{Proof}. Assuming $\mu \in V(\bold{x}_{r_{i}}+\bold{x}_{g_{i}})$ and $\delta = \min_{\lambda \in V({\bold{x}_{r}})}\propto \left | V-\mu \right |=\left ( \left \|(\mu I-V)^{-1} \right \|_{2}) \right )^{-1}$ for identity matrix $I$ and pencil $\left ( \mu I- V \right ) $. According to the perturbation and inverse theorem for nonsingular (or quasi-singular) matrices \cite{golub2012matrix}, for $\delta > 0$: $I-\left ( \mu I- V\right )^{-1}\bold{x}_{g}$ is singular and doubly bounded as:
\begin{equation}
    \mathrm{const}\leq \left \| \left ( \mu I- \bold{x}_{r}\right )^{-1}\bold{x}_{g} \right \|_{F}\leq \left \| \left ( \mu I- \bold{x}_{r}\right )^{-1} \right \|_{F}\left \| \bold{x}_{g} \right \|_{F}
\end{equation}
\noindent where $\mathrm{const}$ is a positive constant scalar and $\left ( \mu I- V\right )^{-1}$ is diagonal in the upper bound of $p_{r}$ and with the assumption of $\left | S \right |^{d}=0 $ it can be approximated by:
\begin{equation}
    \left [ \left ( \mu I- V\right )-S \right ]^{-1}\approx\sum_{k=0}^{d-1}\left [  \left ( \mu I- V\right )^{-1}S^{H}\right ]^{k}\left ( \mu I- V\right )^{-1}
\end{equation}
\noindent which reduces the pencil of $\left ( \mu I- V\right )$ to a small scalar as computed in (\ref{pencil_reduce}) regardless the choice of $Q$ or $Z$ \cite{golub2012matrix}. Consequently, DFN is independent of the variant choices of these basis vectors while it can be used as metric for measuring similarity between subsets of $p_{r}$ and $p_{g}$.
\begin{equation}
    \left \| \left [  \left ( \mu I- V\right ) -S\right ]^{-1} \right \|_{F}\leq \delta^{-1}\left ( \sum_{k=0}^{d-1}\left \| S^{H} \right \|_{F} \right )^{k} \quad \square
    \label{pencil_reduce}
\end{equation}
The generator in a typical GAN configuration unnecessarily learns latent properties of sample distribution (those features of $\bold{x}_{r_{i}}$ and $\bold{x}_{g_{i}}$ encoded in $Q$ and $Z$) which might have convex behavior in the spans of their subspaces, thus increasing the total number of parameters ($\theta_{g}$) in $G$ does not explicitly help to achieve more distinguishable modes; thus it results to collapses against $p_{r}$ distribution with the JSD metric.

%%%%%%%%%%%%%%%%%%%%%%%%%%%%%%%%%%%%%%%%%%%%%
\subsection{DFN Metric for LS-GAN}
%%%%%%%%%%%%%%%%%%%%%%%%%%%%%%%%%%%%%%%%%%%%%
We integrate DFN metric into the generator's optimization formulation of LS-GAN as shown in (\ref{lsgna_d2}) still with the assumption of optimizing toward achieving the optimal discriminator model $\left ( bp_{r}+ap_{g} \right )/p_{r}+p_{g}$ as shown in (\ref{lsgna_d3}) \cite{mao2017least}. We also set the hyperparameters $a$, $b$, and $c$ as suggested in the default configuration of LS-GAN.
\begin{equation}
    \min_{G}\frac{1}{2}\mathop{{}\mathbb{E}}_{\bold{z}\sim p_{z}}\left [ (D(G(\bold{z}))-c)^{2} \right ] \quad \mathrm{s.t.}\quad \left | \mathbb{E}_{\bold{z}\sim p_{z}} \Delta^{2}(G(\bold{z})) - \mathbb{E}_{\bold{x}\sim p_{r}} \Delta^{2}(\bold{x}) \right |<\epsilon
    \label{lsgna_d3}
\end{equation}
\noindent where $\Delta^{2}(.)$ has been defined in (\ref{f_gan2}). Whereas the Wasserstein metric, DFN does not require to support Lipschitz continuity. However, finding the optimal threshold value for $\epsilon$ can be very challenging. Since $\Delta^{2}(.)$ is differentiable in its designated subspaces, we can find an upper bound for $\epsilon$. For all $x:=\max (V)$ we assume $g_{1}(x)=\Delta^{2}(\bold{x}) \in \mathbb{C}^{n+1}$ with degree $n+1$ and $g_{2}(x)=\Delta^{2}(G(\bold{z}))$ in degree of $n$ are differentiable over the interval $\left [ \alpha, \beta \right ]$, therefore we can approximate the error function as the following \cite{phillips2003interpolation}.
%\aktd{Here you introduced the variable scalar $x$. It is not clear what such a scalar represents. It does not seem to be a component of array $\bold{x}$? Is it?}
%respond: I edited it.
\begin{equation}
   e(x)=g_{1}(x)-g_{2, n}(x)=\frac{g^{(n+1)}_{1}(\xi)}{(n+1)!}\prod_{i=0}^{n}(x-x_{i})
    \label{errorfunc}
\end{equation}
\noindent where $\xi \in (\alpha,\beta)$ with the marginal condition using the second derivative $g(.)$ as of $\left | {g}''_{1}(x) \right |<\varrho$ for $0 \leq \varrho <<1$ we write: 
%\aktd{ Is ${g}''$ the 2nd derivative ? And what is $\varrho$? }
% edited.
\begin{equation}
    g_{1}(x)-g_{2,1}(x)=\left ( x-x_{i} \right )\left ( x-x_{i+1} \right )\frac{g{}''_{1}(\xi)}{2!}
\end{equation}
\noindent In the simplest case by substituting $g_{3}(x)=\left ( x-x_{i} \right )\left ( x-x_{i+1} \right )$ and computing its first derivative as follows.
\begin{equation}
    g{}'_{3}(x)=2x-(x_{i}+x_{i+1})=0, \quad \mathrm{then} \quad g_{3}\left ( \frac{x_{i}+x_{i+1}}{2} \right )=-\left ( \frac{x_{i+1}-x_{i}}{2} \right )^{2}
\end{equation}
\noindent therefore the acceptable error for residual DFN is bounded as $\left | e(x) \right |\leq \frac{g{}''_{1}(\xi)}{8}\left ( x_{i+1}-x_{i} \right )^{2}$. Moreover, $\epsilon$ can be set empirically with respect to the performance of $D$ and $G$ networks on the designated dataset.

\begin{figure}[htpb!]
  \centering
  \includegraphics[width=0.75\textwidth]{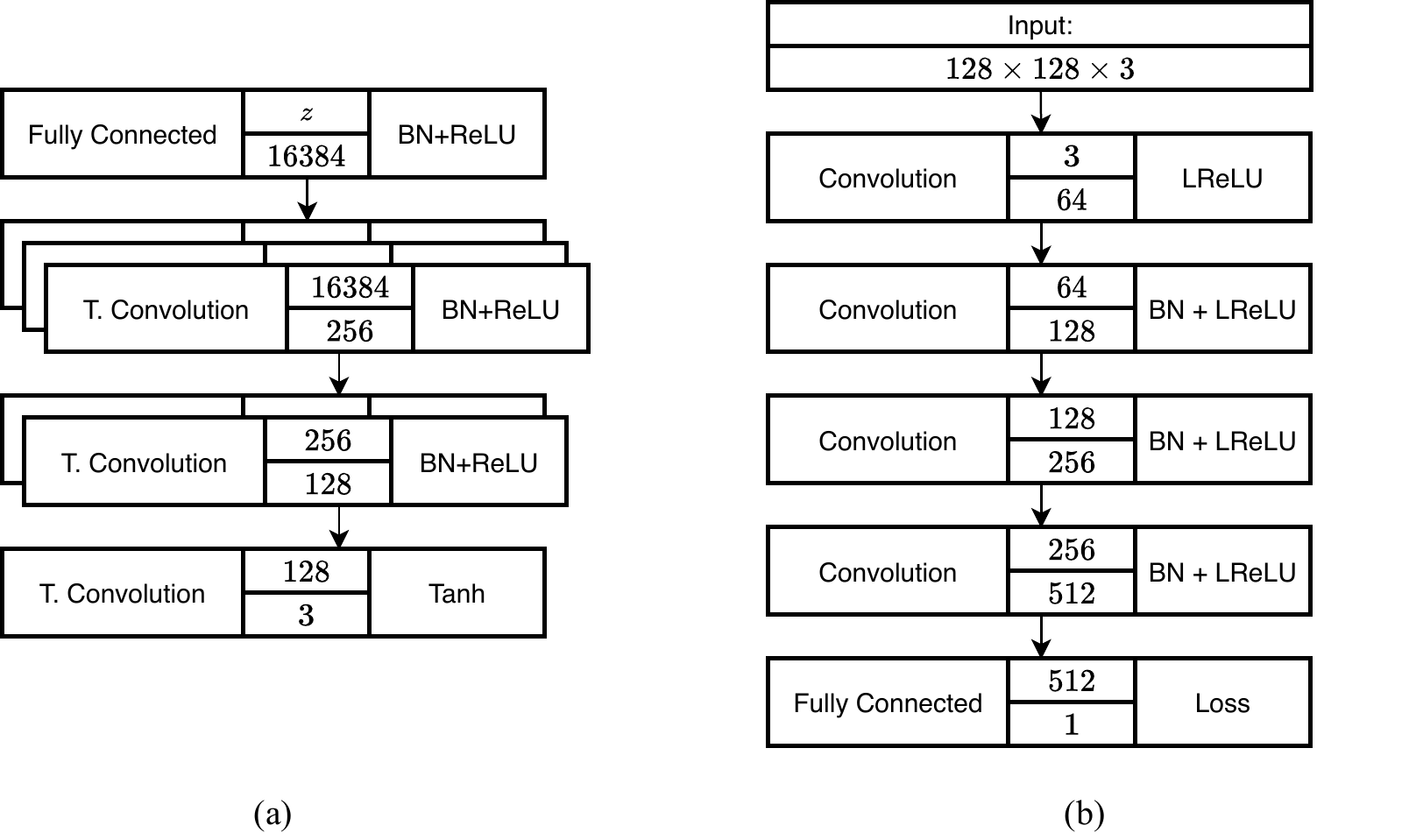}
  %\includepdf{LSGAN_archi.pdf}
  \caption{Architectures for LS-GAN inspired by \cite{mao2018effectiveness}. (a) Generator $G(\bold{z};\theta_{g})$, (b) Discriminator $D(\bold{x}; \theta_{r})$. BN and LReLU stand for batch normalization and Leaky ReLU. Additionally, T. Convolution denotes transposed convolution layers. In the middle blocks, top and bottom values refer to the total number of input and output filters, respectively.}
  \label{LSGAN_archi}
\end{figure}

The intuition behind incorporating DFN into the generator minimization problem is pushing the decision boundary over the possible subspaces of $p_{r}$. This not only might increase the chance of learning more local modes, but also reduces the oversmoothing side effect. For regions in $p_{r}$ where similar local distributions lie in different (but adjacent) subspaces, DFN penalizes $G$ to avoid skipping them. The proposed configurations for the convolutional networks of $G$ and $D$ are shown in Figure~\ref{LSGAN_archi}, and they are inspired in \cite{mao2017least}. This model is designed to fit actual dimensions of waveform representations (DWT spectrograms). For the span of manifolds in Schur subspaces, we square dimensions of all input samples using bilinear interpolation ($128\times 128$). For the generator architecture, all the layers, except for the first and the last, batch normalization has been applied followed by ReLU activation function. For the last transposed convolution layer, $\tanh$ activation functions has been used. Moreover, kernel size has been set to $3\times 3$ with constant stride of $2$. For the convolution layers of the discriminator network, stride is set to $2$ with the kernel size of $5\times 5$ followed by leaky ReLU activation function.  

%%%%%%%%%%%%%%%%%%%%%%%%%%%%%%%%%%%%
\section{Experimental Results}
\label{exp:sec}
%%%%%%%%%%%%%%%%%%%%%%%%%%%%%%%%%%%%
In this section, we carry out some experiments on two benchmarking datasets: UrbanSound8k (US8k) \cite{Salamon:UrbanSound:ACMMM:14} and Mozilla common voice (MCV)\footnote{\url{https://voice.mozilla.org/en/datasets}}. The first dataset includes 8,732 short environmental audio signals ($\leq 4$ sec) organized in 10 different classes. MCV consists of 4,257 recorded hours of multi-language speeches ($\leq 7$ sec) and the corresponding text transcriptions. In our experiments, we randomly select batch of samples from each dataset and generate 2D DWT representations using complex Morlet mother function with static sampling frequency of 16 kHz. Additionally, the frame length has been set up to 50 ms with 50\% overlapping for all recordings. For audio signals with length lower than $2$ sec we apply pitch shifting with scales  $0.75, 0.9, 1.15,$ and $1.5$ as suggested in \cite{esmaeilpour2020detection}. Furthermore, we generate three different visualizations using linear, logarithmic, logarithmic real magnitude scales for spectrogram enhancement purposes \cite{esmaeilpour2019robust}. 

\begin{figure*}[!htpb]
  \subfloat[US8k, $\log$]{\label{fig1:a}\includegraphics[width=0.16\linewidth]{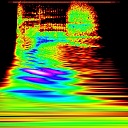}}\hfill
  \subfloat[US8k, $\log Re$]{\label{fig1:b}\includegraphics[width=0.16\linewidth]{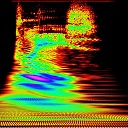}}\hfill
  \subfloat[US8k, linear]{\label{fig1:aa}\includegraphics[width=0.16\linewidth]{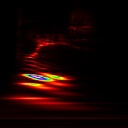}}\hfill
  \subfloat[MCV, linear]{\label{fig1:bb}\includegraphics[width=0.16\linewidth]{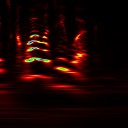}}\hfill
  \subfloat[MCV, $\log Re$]{\label{fig1:aaa}\includegraphics[width=0.16\linewidth]{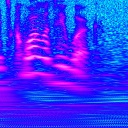}}\hfill
  \subfloat[MVC, $\log$]{\label{fig1:aaaa}\includegraphics[width=0.16\linewidth]{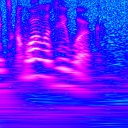}}
  \vspace{-5pt}
  \caption{Generated DWT spectrograms using LS-GAN$_{011}$ with DFN.}
  \label{Cfig}
  \vspace{10pt}
  \subfloat[US8k, $\log$]{\label{fig1:a1}\includegraphics[width=0.16\linewidth]{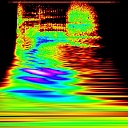}}\hfill
  \subfloat[US8k, $\log Re$]{\label{fig1:b1}\includegraphics[width=0.16\linewidth]{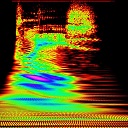}}\hfill
  \subfloat[US8k, linear]{\label{fig1:a2}\includegraphics[width=0.16\linewidth]{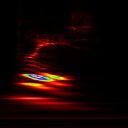}}\hfill
  \subfloat[MCV, linear]{\label{fig1:b2}\includegraphics[width=0.16\linewidth]{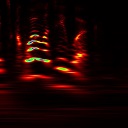}}\hfill
  \subfloat[MCV, $\log Re$]{\label{fig1:a3}\includegraphics[width=0.16\linewidth]{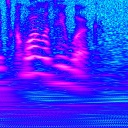}}\hfill
  \subfloat[MVC, $\log$]{\label{fig1:a4}\includegraphics[width=0.16\linewidth]{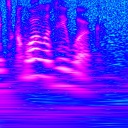}}
    \vspace{-5pt}
  \caption{Generated DWT spectrograms using LS-GAN$_{-110}$ with DFN.}
  \vspace{10pt}
  \subfloat[US8k, $\log$]{\label{fig1:ab}\includegraphics[width=0.16\linewidth]{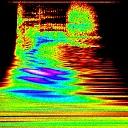}}\hfill
  \subfloat[US8k, $\log Re$]{\label{fig1:bc}\includegraphics[width=0.16\linewidth]{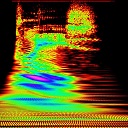}}\hfill
  \subfloat[US8k, linear]{\label{fig1:ac}\includegraphics[width=0.16\linewidth]{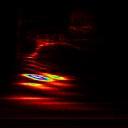}}\hfill
  \subfloat[MCV, linear]{\label{fig1:bd}\includegraphics[width=0.16\linewidth]{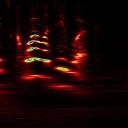}}\hfill
  \subfloat[MCV, $\log Re$]{\label{fig1:ad}\includegraphics[width=0.16\linewidth]{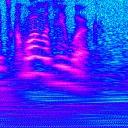}}\hfill
  \subfloat[MVC, $\log$]{\label{fig1:ae}\includegraphics[width=0.16\linewidth]{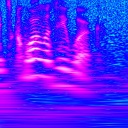}}
    \vspace{-5pt}
  \caption{Generated DWT spectrograms using regular LS-GAN$_{011}$.}
    \vspace{10pt}
  \subfloat[US8k, $\log$]{\label{fig1:a9}\includegraphics[width=0.16\linewidth]{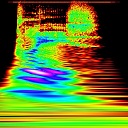}}\hfill
  \subfloat[US8k, $\log Re$]{\label{fig1:b9}\includegraphics[width=0.16\linewidth]{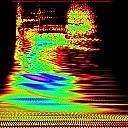}}\hfill
  \subfloat[US8k, linear]{\label{fig1:a99}\includegraphics[width=0.16\linewidth]{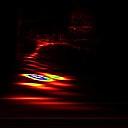}}\hfill
  \subfloat[MCV, linear]{\label{fig1:b99}\includegraphics[width=0.16\linewidth]{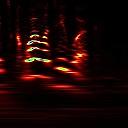}}\hfill
  \subfloat[MCV, $\log Re$]{\label{fig1:a999}\includegraphics[width=0.16\linewidth]{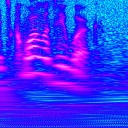}}\hfill
  \subfloat[MVC, $\log$]{\label{fig1:a9999}\includegraphics[width=0.16\linewidth]{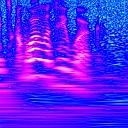}}
    \vspace{-5pt}
  \caption{Generated DWT spectrograms using regular LS-GAN$_{-110}$.}
  \label{Ffig}
\end{figure*}

Visual quality of the generated samples for two random audio signals (one from Ur8k and another from MCV) by generative models including two baseline GANs and two improved GANs are shown in Figures~\ref{Cfig} to \ref{Ffig}. In all these figures, logarithmic and logarithmic real visualizations are denoted by $\log$ and $\log Re$ symbols, respectively. For the generative models, we separately trained the regular LS-GAN with hyperparameters $a=0$, $b=c=1$ (LS-GAN$_{011}$); and $a=-1$, $b=1$, $c=0$ (LS-GAN$_{-110}$). Additionally, we individually trained the aforementioned LS-GANs with these setups using DFN metric encoded in (\ref{lsgna_d3}). All these generative models have been trained using the same architecture for both the generator and the discriminator as shown in Figure~\ref{LSGAN_archi}. For the front-end model selection, we saved multiple model checkpoints for batches of generated spectrograms in every 500 iterations. Finally, candidate models which have been able to produce high quality samples have been selected to generate new spectrograms (see Figures~\ref{Cfig} to~\ref{Ffig}). Since the nature of spectrograms is fundamentally different from natural images, qualitative comparison of generated spectrograms may not reflect a sensible contrast. However, it is obvious that LS-GANs with DFN has generated less noisy and jittery frequency plots compared to baseline GANs.

For quantitatively comparing the performance of the generators in spectrogram production, we conduct two experiments. Firstly, we compute the common Fr\'{e}chet inception distance (FID) metric for evaluating the distance between real and generated samples \cite{heusel2017gans}. Closer distance between samples in their feature spaces should yield smaller FID values. We compute FID values for 2,000 randomly generated spectrograms in every 2,500 iterations. Table~\ref{FID_cmp} shows that both improved LS-GANs outperform baseline GANs, except for the logarithmic real representations of US8k dataset. However, they show competitive performance for the majority of the cases. Moreover, FID values for MCV representations are larger than US8k, on average. We hypothesize that this is due to the structural difference among frequency plots of environmental sounds and speech recordings.

\begin{table}[htpb!]
\centering
\caption{FID values for three spectrogram visualizations (linear and nonlinear) crafted from UrbanSound8k (US8k) and Mozilla Common Voice (MCV) datasets. The best results are in bold.}
\begin{tabular}{|c||c|c|c||c|c|c|}
\hline
                         & \multicolumn{3}{c||}{US8k Representations}   & \multicolumn{3}{c|}{MCV Representations}    \\ \hline
GAN Type                     & linear & $\log$ & $\log Re$ & linear & $\log$ & $\log Re$ \\ \hline \hline
LS-GAN$_{011}$           & 25.17  & 18.33  & 22.08     & 34.89  & 43.33  & 41.58     \\ \hline
LS-GAN$_{-110}$          & 24.29  & 17.82  & \textbf{19.11}     & 30.13  & 37.01  & 35.64     \\ \hline
LS-GAN$_{011}$ with DFN  & 24.07  & 15.45  & 22.73     & 31.46  & 34.77  & \textbf{28.71}     \\ \hline
LS-GAN$_{-110}$ with DFN & \textbf{22.11}  & \textbf{12.47}  & 19.53     & \textbf{29.09}  & \textbf{31.83}  & 32.18     \\ \hline
\end{tabular}
\label{FID_cmp}
\end{table}

Secondly, we compute signal to noise ratio (SNR) for quantitatively measuring discrepancies among generated samples from different LS-GANs with respect to the original signals in 1D Cartesian space, as shown in (\ref{snr}) \cite{kereliuk2015deep}. Towards this end, we reconstruct signals from the generated frequency-plot magnitudes ($\bold{x}_{g_{i}}$) and the prior phase information derived from real samples. For a fair comparison, we also reconstruct original signals from the spectrogram ($\bold{\check{x}}_{r_{i}}$) to remove potential side effects of the complex Morlet transformation.
\begin{equation}
    \mathrm{SNR}_{dB}(\bold{\check{x}}_{r_{i}},\bold{x}_{g_{i}})=20\log_{10}\frac{Pw(\bold{\check{x}}_{r_{i}})}{Pw(\bold{x}_{g_{i}})}
    \label{snr}
\end{equation}
\noindent where $Pw(.)$ denotes the power of the signal. Table~\ref{SNR_cmp} summarizes average SNR ratios for 2,000 samples randomly generated using the four aforementioned LS-GANs. Higher SNR means the reconstructed signal has lower noise. Table~\ref{SNR_cmp} shows that LS-GAN$_{-110}$ with DFN outperforms the other GANs for the majority of the cases.
\begin{table}[htpb!]
\centering
\caption{Average SNR comparisons for generated spectrograms. Higher values are shown in boldface.}
\begin{tabular}{|c||c|c|c||c|c|c|}
\hline
                         & \multicolumn{3}{c||}{US8k Representations}   & \multicolumn{3}{c|}{MCV Representations}    \\ \hline
GANs                     & linear & $\log$ & $\log Re$ & linear & $\log$ & $\log Re$ \\ \hline \hline
LS-GAN$_{011}$           & 44.18  & 33.51  & \textbf{39.22} & 23.19 & 45.13  &  39.18    \\ \hline
LS-GAN$_{-110}$          & 45.91  & 32.16  & 35.07 &  26.81 & \textbf{48.96} &   41.77  \\ \hline
LS-GAN$_{011}$ with DFN  &  51.39 &  33.81 & 32.70& 31.05  & 47.30 &    48.05 \\ \hline
LS-GAN$_{-110}$ with DFN & \textbf{51.64} & \textbf{34.26} & 34.49 & \textbf{35.27} & 44.92 &   \textbf{49.62} \\ \hline
\end{tabular}
\label{SNR_cmp}
\end{table}

FID and SNR are already reliable indicators that the proposed LS-GANs are more stable than baseline GANs. Besides that, we run experiments in two aspects excluding gradient penalty due to its massive computational overhead \cite{gulrajani2017improved}. We trained all the four generative models on the combination of ten Gaussian mixture distributions \cite{metz2016unrolled}. For models that suffer from mode collapse, they generated samples only around few mode(s) disregarding other joint distributions. We observed that baseline GANs started to show extreme mode collapse at around 9,000 iterations, detecting six modes on average. The improved LS-GAN models showed more robustness with detecting eight out of 10 modes. We executed these experiments 100 times per model and counted the total number of generated samples around every two modes. We observed that LS-GAN$_{-110}$ with DFN generates fewer samples compared to other GANs, averaged over three visualizations and two datasets. Lower number of generated samples around modes indicates higher stability and semantic learning. Another approach for determining stability of a generative model is training on a dataset with low variability. US8k compared to MCV has lower variability domain. However, the proposed LS-GANs could finely generate samples with limited number of mode collapse (three on average) on such a dataset. 

%%%%%%%%%%%%%%%%%%%%%%%%%%%%%%%%%%%%%%%%%%%%%%%%%%%%%%%%%%%%%%%%%
\section{Conclusion}
%%%%%%%%%%%%%%%%%%%%%%%%%%%%%%%%%%%%%%%%%%%%%%%%%%%%%%%%%%%%%%%%%
In this paper, we introduced a new similarity metric, the departure from normality, in the unitary vector space of Schur decomposition. We showed that, basis vectors of input samples encode structural components of the sample distribution and since they are independent of eigenvalues, their coefficients can be encoded into the generator network to penalize against mode collapse and improve training stability.

We encoded the DFN metric into the generator network of two regular LS-GANs and also proposed a stable architecture for DWT representation of audio and speech signals. We compared the generated samples of our improved LS-GANs against baseline GANs both qualitatively and quantitatively. Subjective evaluations have shown that less noisy spectrograms can be generated by DFN-based LS-GANs. For quantitative comparisons, we separately measured Fr\'{e}chet inception distance and SNR metrics. Upon conducting several experiments, we concluded that, the proposed LS-GAN setups outperformed other LS-GANs. 

%%%%%%%%%%%%%%%%%%%%%%%%%%%%%%%%%%%%
\section*{Broader Impact}
%%%%%%%%%%%%%%%%%%%%%%%%%%%%%%%%%%%%
This research work proposes an approach for enhancing stability of typical GANs including the state-of-the-art LS-GAN. Although our case study in this paper has been defined in the scope audio and speech signals, it is generalizable to other applications such as computer vision. Due to the nonlinearity characteristics of the convolutional neural network based generative models, it is very difficult to establish a generalizable while stable architecture for GANs. However, this study opens up a new direction for exploring algebraic-based subspaces for developing more robust generative models.  
%%%%%%%%%%%%%%%%%%%%%%%%%%%%%%%%%%%%

%\section*{References}

%References follow the acknowledgments. Use unnumbered first-level heading for
%the references. Any choice of citation style is acceptable as long as you are
%%consistent. It is permissible to reduce the font size to \verb+small+ (9 point)
%when listing the references.
%{\bf Note that the Reference section does not count towards the eight pages of content that are allowed.}
\medskip

\small
\bibliographystyle{IEEEtran}
%\bibliography{mybib}

\end{document}